\UseRawInputEncoding
\documentclass{article}
\usepackage[preprint]{colm2025_conference}

\usepackage{microtype}
\usepackage{hyperref}
\usepackage{url}
\usepackage{booktabs}
\usepackage{graphicx}
\usepackage{multirow}
\usepackage{subcaption}
\usepackage{amsmath,amssymb,amsfonts}
\usepackage{xcolor}
\usepackage{lineno}
\usepackage{tcolorbox}
\usepackage{enumitem}
\usepackage{tikz}
\usepackage{listings}
\usetikzlibrary{calc, shapes, arrows.meta, positioning, shadows}

\definecolor{darkblue}{rgb}{0, 0, 0.5}
\hypersetup{colorlinks=true, citecolor=darkblue, linkcolor=darkblue, urlcolor=darkblue}

\author{%
Charles O'Neill \\
University of Oxford \\
\texttt{cponeill00@gmail.com}
\And
Tirthankar Ghosal \\
Oak Ridge National Laboratory \\
\texttt{ghosalt@ornl.gov}
\And
Roberta R\u{a}ileanu \\
University College London \\
\texttt{r.raileanu@ucl.ac.uk}
\And
Mike Walmsley \\
University of Toronto \\
\texttt{m.walmsley@utoronto.ca}
\And
Thang Bui \\
Australian National University \\
\texttt{thang.bui@anu.edu.au}
\And
Kevin Schawinski \\
Modulos AG \\
\texttt{schawinski@gmail.com}
\And
Ioana Ciuc\u{a} \\
Stanford University \\
\texttt{iciuca@stanford.edu}
}

\title{Sparks of Science: Hypothesis Generation Using Structured Paper Data}

\begin{document}

\ifcolmsubmission
\linenumbers
\fi

\maketitle

\begin{abstract}
Generating novel and creative scientific hypotheses is a cornerstone in achieving Artificial General Intelligence. Large language and reasoning models have the potential to aid in the systematic creation, selection, and validation of scientifically informed hypotheses. However, current foundation models often struggle to produce scientific ideas that are both novel and feasible. One reason is the lack of a dedicated dataset that frames Scientific Hypothesis Generation (SHG) as a Natural Language Generation (NLG) task. In this paper, we introduce \textit{HypoGen}, the first dataset of approximately 5500 structured problem-hypothesis pairs extracted from top-tier computer science conferences structured with a \textbf{Bit-Flip-Spark} schema, where the \textbf{Bit} is the conventional assumption, the \textbf{Spark} is the key insight or conceptual leap, and the \textbf{Flip} is the resulting counterproposal. \textit{HypoGen} uniquely integrates an explicit \textbf{Chain-of-Reasoning} component that reflects the intellectual process from \textbf{Bit} to \textbf{Flip}. We demonstrate that framing hypothesis generation as conditional language modelling, with the model fine-tuned on \textbf{Bit-Flip-Spark} and the \textbf{Chain-of-Reasoning} (and where, at inference, we only provide the \textbf{Bit}), leads to improvements in the overall quality of the hypotheses. Our evaluation employs automated metrics and LLM judge rankings for overall quality assessment. We show that by fine-tuning on our \textit{HypoGen} dataset we improve the novelty, feasibility, and overall quality of the generated hypotheses. The \textit{HypoGen} dataset is publicly available at
\href{https://huggingface.co/datasets/UniverseTBD/hypogen-dr1}{huggingface.co/datasets/UniverseTBD/hypogen-dr1}.

\end{abstract}

\section{Introduction}

Hypothesis generation is the first step of the scientific process and its de facto foundation. Creative and innovative ideas have long enabled scientists to model and predict the behaviour of complex systems, from neuroscience to astrophysics. Recently, the impressive capabilities of large language models have prompted researchers to explore their potential to advance the generation of scientific ideas \citep{Ziems2023CanLL, Birhane2023ScienceIT, Xie2023LargeLM,Noever2023NumeracyFL, si2024novel, kumar2024unlock, xiong2024improving, zhou2024hypothesis, cohrs2025large}. Not only do these models excel in understanding and generating human language \cite[e.g.,][]{devlin2018bert,brown2020language,team2023gemini,grattafiori2024llama}, but they also demonstrate a remarkable ability to make nuanced deductions and establish relationships across varied contexts \citep{elkins2020can}, rendering them an ideal basis for the generation of semantic hypotheses. Recent work has evaluated LLMs on the entire scientific discovery process, from hypothesis generation to running experiments, analyzing the results, and even writing a paper~\citep{lu2024ai, chan2024mle,chen2024scienceagentbench, gottweis2025towards,nathani2025mlgym,schmidgall2025agent,schmidgall2025agentrxiv}. However, most works highlight limitations of current models when applied to open research problems, particularly with respect to generating novel, creative, diverse, feasible, actionable, interesting, and useful ideas or hypotheses~\citep{nathani2025mlgym}.

LLMs face significant challenges when applied to scientific ideation. These models are prone to hallucinations, often producing non-factual content due to their token likelihood maximization objective \citep{Manakul2023SelfCheckGPTZB, McKenna2023SourcesOH, Li2023HaluEvalAL, Zhang2023UserControlledKF, tonmoy2024hallucination, lu2024ai}. Recent benchmarks highlight that such inaccuracies can be difficult to detect, as LLMs often present them with high confidence \citep{Qi2023LargeLM, Zhou2024HypothesisGW}. Additionally, probability-maximizing decoding strategies (e.g., greedy or high-beam search) can lead to text that lacks lexical diversity, a problem that persists even in models with hundreds of billions of parameters \citep{Holtzman2019TheCC, Li2022ContrastiveDO, Meister2022TypicalDF, Su2022ACF, Zhou2024HypothesisGW}. 

\begin{figure}[t]
    \centering
    % Include the image and scale it to the text width
    \includegraphics[width=0.9\linewidth]{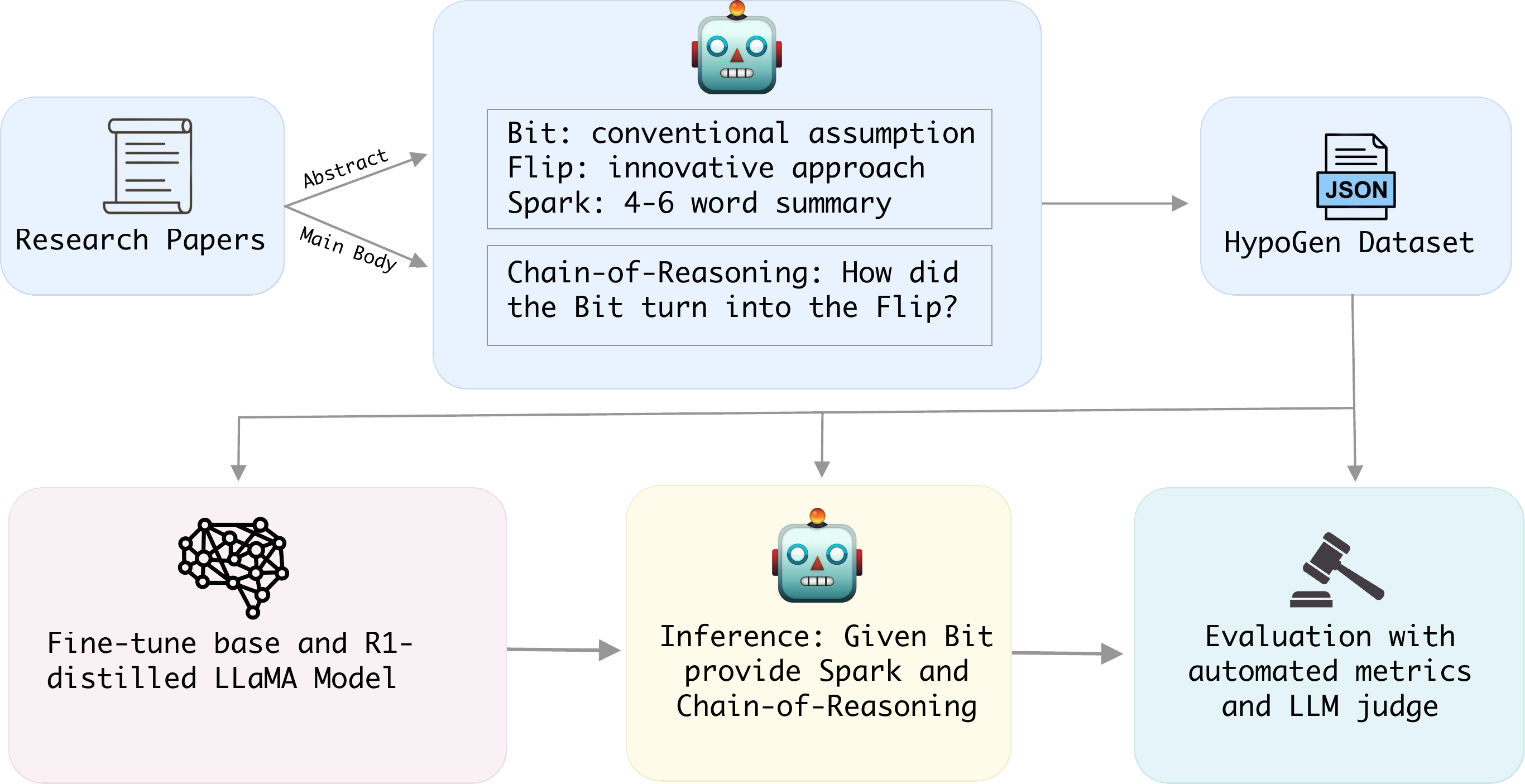}
    \caption{The \textit{HypoGen} process begins with input paper abstracts, from which the structured \textbf{Bit} (the problem), \textbf{Flip} (the solution) and \textbf{Spark} (key insight) are extracted by OpenAI's \texttt{o1} model. The \textbf{Chain of Reasoning} is extracted by the \texttt{o1} model from the main body of the paper. These outputs are used to fine-tune a LLaMA-based model, which then generates hypotheses from the provided \textbf{Bit}. A judge module (\texttt{Claude 3.7 Sonnet}) assesses the overall quality based on novelty and feasibility.}
    \label{fig:pipeline}
\end{figure}

%evaluation is hard
The design of a validation scheme to rigorously test these machine-generated hypotheses poses additional challenges \citep{Alaa2021HowFI, si2024novel, Luo2025LLM4SRAS}. To be effective, scientific hypotheses not only require creative insight drawn from a broad understanding of the domain at hand, but also must be rooted in the existing literature to ensure their novelty and relevance \citep{simonton2004creativity, runco2012standard, doboli2014new, strom2018creativity, Wang2023SciMONSI}. In addition, it is difficult to determine in an automated fashion to what extent a certain idea already exists in the literature, which is particularly problematic due to the tendency of LLMs to copy subsets of their training data in generation \citep{McCoy2021HowMD, Liu2021CanAT}. Given that validation is integral to the scientific method, the closed-box nature of LLMs requires a careful and nuanced approach to ensure that the results are replicable and robust. 

% FLIP THE BIT - our approach
To address these challenges, we introduce \textit{HypoGen}, a dataset comprising approximately 5500 structured problem-hypothesis pairs extracted from top-tier computer science conferences. This dataset represents a significant step forward in framing scientific hypothesis generation as a conditional language modeling problem. By conditioning hypotheses on a clear formulation of the problem (the \textbf{Bit}), our approach provides a robust foundation for developing and evaluating LLMs in the context of scientific discovery. Importantly, \textit{HypoGen} incorporates a detailed \textbf{Chain-of-Reasoning} narrative that mirrors the iterative and reflective process used by human scientists to transition from conventional wisdom to innovative counterproposals, thus improving both the quality and the trustworthiness of the generated hypotheses.

Our key contributions include the development of the \textit{HypoGen} dataset and the novel framing of scientific hypothesis generation as a conditional language modeling problem enriched with an explicit reasoning chain. We present baseline performance measures of an LLaMA-based model on a hypothesis generation task after being fine-tuned on the \textit{HypoGen} dataset. We employ a straightforward evaluation framework that assesses hypotheses along the dimensions of novelty and feasibility, incorporating automated metrics and LLM judgements. By capturing the full chain of reasoning, our approach provides valuable insights into the thought processes underlying scientific discovery.

\section{Related Work}

Several approaches factor the decision process into sub-stages. In the \emph{proposal} stage, reasoning and sometimes retrieval are used to generate candidate actions or hypotheses \citep{chen2021evaluating, wang2022self}. The \emph{evaluation} stage then scores these candidates (for example, perplexity \citep{ahn2022can} or learned reward functions \citep{yao2020keep}), identifying which candidates are the most promising. Techniques such as ToT \citep{yao2023tree} and RAP \citep{hao2023reasoning} use tree search paradigms to propose and evaluate multiple solution paths in a structured manner. Reflexive approaches such as \cite{shinn2023reflexion} and \cite{lindes2023improving} explicitly incorporate iterative self-correction of hypothesized actions. The work of \cite{Zhou2024HypothesisGW} with \emph{HypoGeniC} expands this process with iterative reinforcement learning with human feedback. 

These advances stress the need for benchmarks that realistically reflect the capacity of LLMs to generate, validate, and refine scientific hypotheses \cite[e.g.,][]{kumar2024unlock, Majumder2024DiscoveryBenchTD,Luo2025LLM4SRAS}. For example, the ``Knowledge Grounded Chain of Ideas" or KG-CoI system \citep{Xiong2024ImprovingSH} removes specific links from a biomedical knowledge graph and asks LLMs to propose plausible missing relations. Because these links are derived from previously held information, LLM-generated hypotheses can be validated against known ground truths. Such tasks resemble real-world discovery scenarios, where a laboratory of AI agents can interact with human experts, document interactions, and call tools to achieve a particular task, for example, to design a novel protein binder \cite[e.g.,][]{Swanson2024}. Other innovative evaluation environments, such as \emph{Discovery World} \citep{Jansen2024DISCOVERYWORLDAV} or \emph{AI Scientist} \citep{Lu2024TheAS}, provide virtual environments where an AI agent can \emph{propose} hypotheses and \emph{conduct} simulated experiments, opening the possibility of end-to-end science.

However, there remains a lack of standardized ``frontier" benchmarks designed to evaluate hypothesis generation capabilities, especially in the context of agentic AI systems, which rely on highly interconnected modules that require complex reasoning \citep{Shao2024CollaborativeGA}. To this end, we introduce \textit{HypoGen}, a benchmark dataset specifically designed to address current deficits in the evaluation of the generation of scientific hypotheses. In contrast to existing benchmarks, \textit{HypoGen} explicitly emphasizes \textbf{Chain-of-Reasoning}: each hypothesis includes a transparent \emph{abductive} logic trail that mirrors the thought process of a human expert. Our method uses a structured \textbf{Bit-Flip-Spark + Chain-of-Reasoning} format to capture the conceptual progression from an initial problem statement (\textbf{Bit}), to a key insight (\textbf{Spark}), and finally to a refined idea (\textbf{Flip}). By incorporating a detailed reasoning chain, \textit{HypoGen} helps mitigate the risk of hallucination \citep{tonmoy2024hallucination}, while simultaneously providing researchers with a reproducible step-by-step notebook of \emph{how} a new idea was generated.

\section{Methodology and the \textbf{Bit-Flip-Spark+Chain-of-Reasoning} Format}
\label{sec:methodology}

Figure~\ref{fig:pipeline} illustrates the overall pipeline of \textit{HypoGen} \footnote{Our code implementation is publicly available at
\href{https://github.com/UniverseTBD/hypogen-cs}{github.com/UniverseTBD/hypogen-cs}.}, which is designed to extract structured information from scientific papers using the \textbf{Bit-Flip-Spark+Chain-of-Reasoning} format. The Stanford Bit-Flip schema\footnote{\href{https://web.stanford.edu/class/cs197c/slides/02-literature-search.pdf}{\texttt{https://web.stanford.edu/class/cs197c/slides/02-literature-search.pdf}}} serves as a concise and structured hypothesis formulation technique designed to encapsulate the core intellectual contribution of a research paper. The \textbf{Bit} identifies the prevailing belief or assumption in the research domain that the paper aims to challenge. The \textbf{Flip} articulates the novel approach or counterargument that the paper introduces to advance the field. In addition, we introduce a novel concept, \textbf{Spark}, which contains the ``essence of an idea", formalized as a conceptual leap. We provide an example extracted from \cite{bahdanau2015neural} and the full prompt to obtain this representation in Appendix \ref{sec:appendix_extras}.

\begin{tcolorbox}[
  colback=teal!10,    % Light, natural pastel teal background
  colframe=teal!70,   % Deeper teal border for an elegant contrast
  title=Example \textit{HypoGen} Hypothesis,
  fonttitle=\bfseries,
  boxrule=1pt,
  arc=4mm,
  auto outer arc
]
\textbf{Bit}: Conventional neural machine translation encodes the entire source sentence into a single fixed-length vector. This approach imposes a bottleneck, because crucial contextual information can be lost in the compression process.\\[1mm]
\textbf{Spark}: Soft alignment for flexible translation.\\[1mm]
\textbf{Flip}: An idea was introduced that enables the model to automatically search for relevant parts of the source sentence at each decoding step. This removes the need for a single fixed-length representation by allowing a soft-alignment mechanism that dynamically focuses on important source information.\\[1mm]
\textbf{Chain-of-Reasoning}: I started by considering that traditionally neural machine translation relies on a single fixed-length vector to encode an entire source sentence before generating the target sequence, and that this rigid bottleneck can lose important contextual information, limiting translation quality and flexibility. I questioned whether a more dynamic way to capture contextual cues was needed, especially for longer sequences. Instead, I introduced a learnable (soft) search mechanism that dynamically aligns each target word with the most relevant parts of the source sentence, shifting away from a fixed-length encoding to attend to context more flexibly and generate higher-quality translations. I trained a bidirectional encoder–decoder on a large bilingual corpus, used a thirty-thousand-word vocabulary, and optimized it via minibatch gradient methods while testing different hidden-layer sizes. How did we confirm that the new approach was identifying correct alignments? I visualized alignment weights and observed that the decoder selectively focused on relevant source words, substantially improving clarity and accuracy. A follow-up question examined how to handle unknown or rare terms, prompting an exploration of further lexical coverage strategies. I recognized a turning point when it became evident that soft attention preserved essential details for both short and long inputs. I validated performance on held-out test sets, noting that the model equaled or surpassed phrase-based benchmarks and maintained robustness on lengthy sentences. This integrated reasoning closed the gap between the original limitation and the dynamic alignment concept, paving the way for more context-aware neural translation.
\end{tcolorbox}

The objective is to distill the complex ideas within a paper into a simplified yet rigorous representation, allowing for clear communication of both the problem being tackled (\textbf{Flip}) and the proposed solution (\textbf{Bit}). This approach is grounded in the understanding that a well-articulated hypothesis is the cornerstone of impactful research. Although this structured representation of hypotheses is subjective and is merely one of many options, we found that it worked well for the generation of a solution (i.e.  \textbf{Flip}) conditioned on a problem (i.e., the \textbf{Bit}). Finally, the \textbf{Chain-of-Reasoning} presents a detailed narrative that captures the scientist's ideation process that connects the Bit to the Flip.

\subsection{Preprocessing and Dataset Construction}
We compile our dataset from papers accepted at the two top-tier computer science conferences, NeurIPS 2023 (3218 papers) and ICLR 2024 (2260 papers), resulting in 5478 distinct samples. We then used OpenAI's \texttt{o1} model for the structured extraction step. For each paper, we first extract the \textbf{Bit}, \textbf{Flip}, and \textbf{Spark} components from the abstract. We prompted \texttt{ o1} to identify the conventional assumption, the innovative approach, and a concise 4-6-word summary of the core insight. We then used a robust parallel processing approach with a retry mechanism with up to three attempts per extraction to ensure high-quality output.

For papers with available full text, we extract the \textbf{Chain-of-Reasoning} component using a separate prompt that guides the model to recreate the intellectual progression from \textbf{Bit} to \textbf{Flip}. This step removes the abstract section from the full text to prevent redundancy. It then processes the paper to generate a first-person narrative detailing the scientist's ideation process. We store the output in JSON format and include metadata such as the paper ID, title, authors, venue, year, and citation information. We construct an independent test set of 50 hypotheses from the authors' recent submissions and relevant work between 2024 and 2025.

\subsection{Fine-tuning and Inference Pipeline}
Our baseline models include Meta LLaMA 3.1 8B and R1-distilled LLaMA 3.1 8B. These models are trained on extensive corpora with a context window of 128,000 tokens and employ byte-pair encoding for tokenization \citep{Sennrich2015NeuralMT,kudo2018sentencepiece}, incorporating a vocabulary of 128,000 tokens. The R1-distilled LLaMA 3.1 8B is a specialized model with knowledge transferred from the larger DeepSeek-R1 model with 671B parameters. This substantial pre-training provides robust language understanding capabilities essential for scientific hypothesis generation.

We leverage our curated dataset of structured problem-hypothesis pairs for fine-tuning, employing the causal language modeling objective. The process utilizes four NVIDIA H100 GPUs, each with 80GB of VRAM. We implement 4-bit quantization and deploy LoRA \citep{hu2021lora} with hyperparameters: $\alpha=16$ and a dropout rate of 0.1. The models are loaded with 4-bit precision base loading, using appropriate compute precision (bf16 where supported otherwise fp16). We use the AdamW 8-bit optimizer \citep{loshchilov2017decoupled} with a weight decay of 0.01, a batch size of 32, and a learning rate of $2\times 10^{-4}$. The training follows a linear scheduler with 5 warmup steps and proceeds for approximately 60 total steps, with logging at each step. During inference, only the \textbf{Bit} is provided to the model. The model then generates the corresponding \textbf{Spark} along with a detailed \textbf{Chain-of-Reasoning}. We use the \texttt{ollama} LLM framework for the LLaMA one-shot inference. 

\section{Evaluation}

The task of evaluating generative models tailored for the generation of scientific hypotheses is challenging, given the inherently subjective nature of scientific research. In this paper, we focus on a dual evaluation framework that primarily incorporates traditional automated metrics and LLM-based judges. 

Our evaluation strategy relies on a test set of 50 hypotheses extracted from the recent literature from primarily 2024 and 2025. It combines automated metrics with an LLM Judge module that assesses novelty, feasibility, and overall quality from pairwise comparisons. We further test the robustness of our approach with a second LLM judge. For a subset of our evaluation set, we also use human evaluation to assess whether fine-tuning LLaMA-base models on our \textit{HypoGen} dataset improves the quality of the hypotheses.

\paragraph{Automated Evaluation Metrics} \textit{Perplexity} is used as a preliminary metric to assess the fluency and coherence of the hypotheses generated \citep{chen1998evaluation}. It is defined as the exponentiated average negative log-likelihood of a given token sequence $X=(x_0, x_1, \ldots x_t)$. Mathematically, this is expressed as:
\begin{equation}
\mathrm{PPL}(X)=\exp \left\{-\frac{1}{t} \sum_i^t \log p_\theta\left(x_i \mid x_{<i}\right)\right\}
\end{equation}
Here, $\log p_\theta (x_i | x_{<i})$ denotes the log-likelihood of the $i$-th token conditioned on its preceding tokens according to the model. The metric serves as an indicator of the predictive performance of the model, with lower values suggesting better generalization.

\textit{IAScore} quantifies alignment between LLM-generated hypotheses and expert-proposed research ideas. For each paper $j$, the IAScore computes the average alignment between author-proposed future research ideas (AP-FRI$_j$) and each generated idea $I_{ij}$ using an IdeaMatcher (IM) model \citep{kumar2024unlock}:
\begin{equation}
\text{AvgScore}_j = \frac{1}{N_j} \sum_{i=1}^{N_j} \text{IM}(\text{AP-FRI}_j, I_{ij})
\end{equation}
The domain-wide IAScore for model $M$ is then calculated by averaging across all $P$ papers:
\begin{equation}
\text{IAScore}_{domain,M} = \frac{1}{P} \sum_{j=1}^{P} \text{AvgScore}_j
\end{equation}
\cite{kumar2024unlock} employed GPT as the IdeaMatcher due to its superior performance (91.8\% accuracy) compared to Natural Language Inference using RoBERTa MNLI and BERTScore in determining if a generated idea is contained within the author's proposals. Higher IAScore values indicate greater alignment between LLM-generated ideas and author perspectives across the domain. 

\textit{Idea Distinctiveness Index} evaluates the semantic diversity between the hypotheses generated using embedding-based similarity rather than textual differences at the surface level. For a set of ideas $I$, each idea $id_i$ is embedded into vector $v_i$ using a pre-trained BERT model \citep{kumar2024unlock}. The distinctness between ideas $id_i$ and $id_j$ is defined as $D_{ij} = 1 - \text{sim}(v_i, v_j)$, where $\text{sim}$ is cosine similarity. The overall distinctiveness for a set of $n$ ideas is:
\begin{equation}
\text{D}_I = \frac{1}{n(n-1)} \sum_{i=1}^{n}\sum_{\substack{j=1 \\ j \neq i}}^{n} D_{ij}
\end{equation}
To assess the performance of a model within a domain, we can calculate the Idea Distinctness Index $\text{D}_{Ip_M}$ for all ideas generated by model $M$ for each paper $p$, then average across all $m$ papers:
\begin{equation}
D_{domain,M} = \frac{1}{m} \sum_{p=1}^{m} \text{D}_{Ip_M}
\end{equation}
Higher $D_{domain,M}$ values signify greater idea diversity, indicating the model's ability to generate semantically varied hypotheses within the domain.

\begin{table}[t]
  \centering
  \footnotesize
  \begin{tabular}{l|c|c|c}
    \toprule
    \textbf{Model} & \textbf{Perplexity} & \textbf{IAScore} & \textbf{Idea Distinctness Index} \\
    \midrule
    Gold Outputs & 89.31 & -- & -- \\
    \midrule
    \multicolumn{4}{c}{\textbf{Before Finetuning}} \\
    \midrule
    LLaMA 3.1 8B & 16.70 & 0.2781 & 0.6669 \\
    R1 Distilled LlaMA 3.1 8B & 19.85 & 0.6049 & 0.7146 \\
    \midrule
    \multicolumn{4}{c}{\textbf{After Finetuning}} \\
    \midrule
    LLaMA 3.1 8B - FT & 32.41 & 0.6746 & 0.6297 \\
    R1 Distilled LlaMA 3.1 8B - FT & 34.98 & 0.6729 & 0.6288 \\
    \bottomrule
  \end{tabular}
  \caption{Automated evaluation metrics comparing different model outputs. IAScore measures idea alignment with source material, while the Idea Distinctness Index quantifies the uniqueness of generated hypotheses.}
  \label{tab:automated_metrics}
\end{table}

\paragraph{LLM Evaluation}

To evaluate the quality of the hypotheses in our evaluation set, we employed Anthropic's \texttt{Claude 3.7 Sonnet-Thinking} model as the automated evaluator. We perform a pairwise evaluation on each dataset consisting of 50 problems and proposals of paired solutions generated by two LLMs for each evaluation experiment. We have nine experiments corresponding to
LLaMA 3.1-8B-FT (LLaMA-8B-FT for brevity) vs Human, LlaMA 3.1-8B-FT (LLaMA-8B-FT) vs an \texttt{o1} model with one example (1shot), followed by an R1-distilled-LlaMA-3.1-8B-FT (R1-distilled-LlaMA-FT) vs Human and \texttt{o1}-1shot, LLaMA-8b-FT vs R1-distilled-LLaMA-8b-FT, Human vs \texttt{o1}-1shot, R1-distilled-LlaMA-8b-1shot vs R1-distilled-LLaMA-8b-FT, LLaMA-8B-1shot vs LLaMA-8B-FT and LLaMA-8B-1shot vs R1-distilled-LLaMA-8B-1shot (R1-distilled-LlaMA-1shot). We provide our results in Fig. \ref{fig:results}. The Human hypotheses are the \texttt{o1} structured hypotheses generated from the evaluation set.

For each \textbf{Bit}, the LLM evaluator was asked to evaluate which proposal (\textbf{Spark} + \textbf{Chain-of-Reasoning}) provided the overall better proposal, taking into account novelty and feasibility. We randomize the presentation order of the solutions to mitigate order effects. After each evaluation experiment, we obtain whether proposal A wins in novelty, feasibility, and overall, with an option for a tie. The model's ``thinking'' is further enabled with an 8,000 token budget to encourage thorough deliberation. 

The LLM-based evaluation provides consistency and scalability; however, it comes at the cost of robustness and verifiability. To account for some of these challenges, we rerun our experimental analysis with the OpenAI \texttt{o3-mini} model as a judge to see the degree of agreement. In addition, we conducted a blind human evaluation with 20 hypothesis pairs evaluated by one of the authors. We provide our complete prompts in Appendix \ref{sec:appendix_extras}. 

\section{Results}

\paragraph{Results from Automated Metrics} Table \ref{tab:automated_metrics} shows that human-generated hypotheses have much higher perplexity values than their LLM counterparts. In particular, LLaMA base models exhibit values between 16.70 and 34.98 compared to human ones (89.31). This could point to the semantic creativity present in human-generated ideas. Although perplexity remains lower overall, fine-tuning increases the perplexity score of the LLaMA models, indicating increased ``unpredictability" as it stands to ideation. 

\begin{figure}[t]
    \centering
    % Include the image and scale it to the text width
    \includegraphics[width=\linewidth]{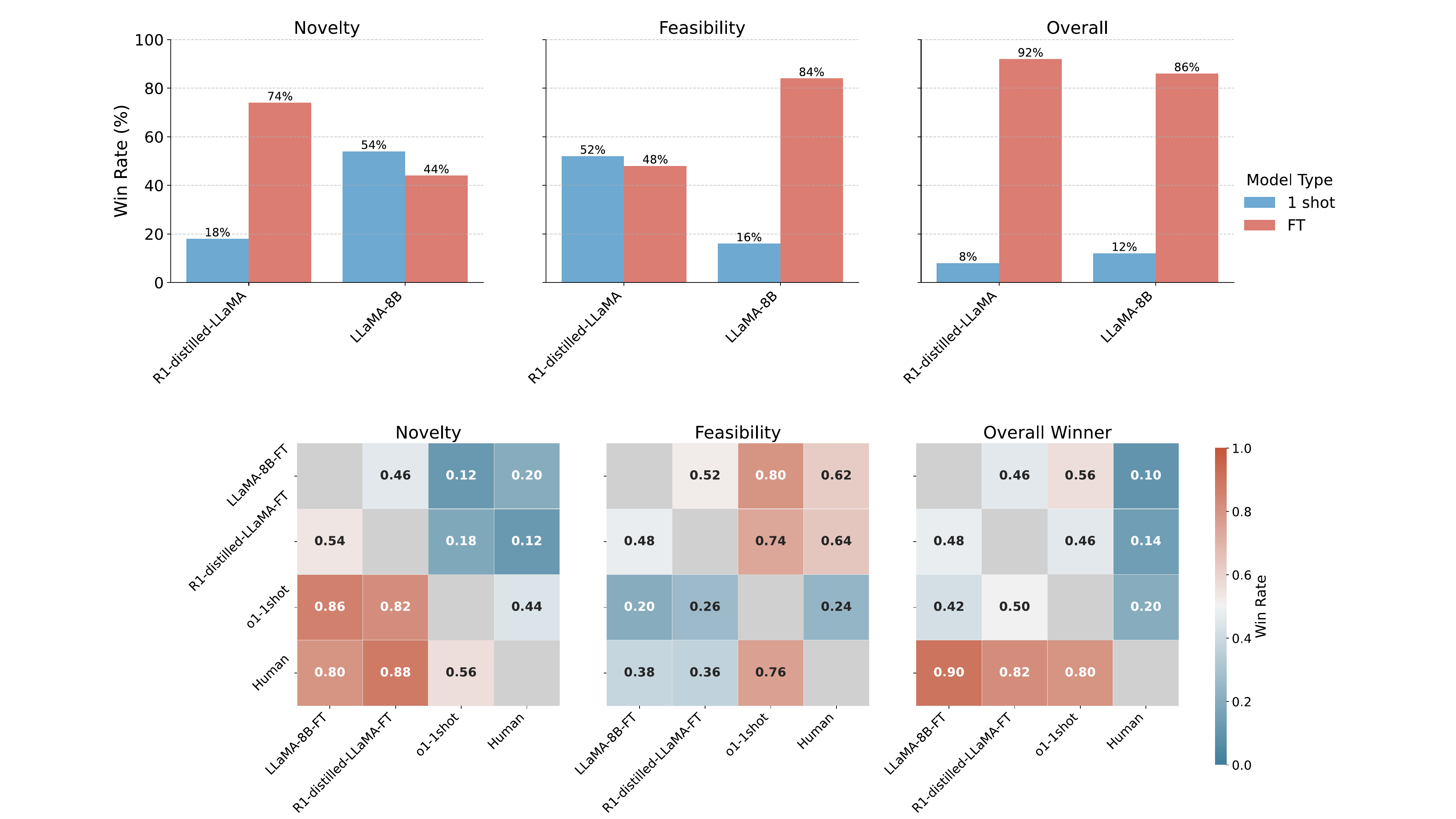}
    \caption{Comparative analysis of the quality of generated hypotheses across nine experiments as evaluated by an LLM Judge \texttt{Claude 3.7 Sonnet}. \textbf{Upper}: Win rates comparing non-fine-tuned versus fine-tuned LLaMA 3.1-8B (LlaMA-8B-FT) and R1-distilled-LlaMA-3.1-8B (R1-distilled-8B-FT) models on novelty and feasibility, showing the consistent trade-off in which fine-tuned models excel at feasibility (74-86\% win rate). Non-fine-tuned variants show greater novelty (54-86\% win rate). \textbf{Lower}: Pairwise win rate heatmap (read on the horizontal) between human experts, fine-tuned models (LLaMA-8B-FT, R1-FT), and one-shot models (O1-1shot, LLaMA-8B-1shot, R1-1shot) across novelty, feasibility, and overall quality dimensions. Human hypotheses are the overall winners (82-90\% win rate), with fine-tuned models achieving comparable feasibility scores (62-64\% vs Human). The fine-tuned models perform better than their one-shot counterparts in overall quality (86-92\% win rate).}
    \label{fig:results}
\end{figure}

Secondly, fine-tuning improves idea alignment with the target domain, as shown by the significant improvement in IAScore for the standard LLaMA model (0.2781 $\rightarrow$ 0.6746). This result could mean that the structured \textbf{Bit-Flip-Spark}+\textbf{Chain-of-Reasoning} training enables models to generate hypotheses that better align with expert-level scientific thinking. The fact that we do not see this effect to the same extent in the distilled LLaMA model may hint at the effectiveness of knowledge transfer.
  
The inverse relationship between IAScore improvements and Idea Distinctness Index reductions, which are particularly notable in the R1 Distilled LLaMA with a reduction from 0.7146 $\rightarrow$ 0.6288, indicates a possible trade-off in hypothesis generation: as models better align with expert scientific thinking patterns, they may produce less semantically diverse outputs. 

\paragraph{Pairwise Comparison using LLM Judges} As shown in the upper panel of Fig. \ref{fig:results}, fine-tuning consistently improves overall hypothesis quality relative to one-shot variants of the same architecture (86-92\% preference for fine-tuned versions), despite the reduction in novelty scores. This indicates that fine-tuning on \textit{HypoGen} steers models toward generating more practical hypotheses.

The LLM evaluation results in \ref{fig:results} reveal a consistent trade-off between novelty and feasibility in the different experiments. Models that excel in creativity metrics seem to underperform in feasibility and vice versa. Human-generated hypotheses win overall in quality assessments compared to LLM-generated alternatives, with human ideas preferred in 80-90\% of the comparisons. However, fine-tuned models demonstrate comparable feasibility scores relative to the human set (A=62-64\% vs. B=36-38\%). Rerunning our analysis with \texttt{o3-mini} as the LLM judge shows consistent behaviour across most experiment: agreement on the key novelty-feasibility trade-off in fine-tuned versus one-shot models and confirming the win of human hypotheses for overall quality. We show our results in Fig. \ref{fig:results-o3} in Appendix \ref{sec:o3-mini}.

\paragraph{Human Evaluation Results}
The results of the small-scale human evaluation trace the observed patterns with the \texttt{Claude 3.7 Sonnet Thinking} model. For the R1-distilled LLaMA comparison, the human evaluator preferred fine-tuned model outputs for novelty (95\% vs. 5\%) and feasibility (70\% vs. 30\%), with an overall preference for fine-tuned outputs (70\% preference, 25\% tie, 5\% base model). The standard LLaMA-8B comparison revealed more competitive performance, with the fine-tuned model maintaining modest advantages in novelty (47. 6\% vs 42. 9\%, 9. 5\% tie) and feasibility (52. 4\% vs 42. 9\%, 4. 8\% tie), resulting in a narrower overall preference (42. 9\% fine-tuned, 33. 3\% one shot, 23. 8\% tie). The human evaluation provides further evidence that fine-tuning on structured \textbf{Bit-Flip-Spark}+\textbf{Chain-of-Reasoning} data improves hypothesis quality, with particularly dramatic improvements observed in the R1-distilled architecture. However, further human evaluation is needed.

\section{Discussion and Future Work}

We introduced the \textit{HypoGen} dataset for the generation of scientific hypothesis that extends the conventional \textbf{Bit-Flip-Spark} format by incorporating a detailed \textbf{Chain-of-Reasoning} component. We showed that fine-tuning on \textit{HypoGen} enables the LLaMA 3.1-8B and R1-distilled-LLaMA 3.1-8B models to improve their hypotheses. This demonstrates the effectiveness of fine-tuning in the intermediate steps of an idea, which provides more transparency and interpretability.  We release \textit{HypoGen} under an MIT license to encourage the development of AI agents capable of supporting human experts in the ideation process.

The primary limitation of \textit{HypoGen} is that it uses LLMs to evaluate the hypotheses generated. Although LLM-as-a-judge modules can perform robustly under certain conditions~\citep{lu2024ai}, they may be biased by their training regime in highly non-trivial ways. To mitigate these unexpected effects, we plan to perform an extensive human evaluation to determine the degree to which human and LLM align on a particular judgement. These findings will guide the construction of more robust reward models that align closely with human expertise, further strengthening \textit{HypoGen}’s applicability in real-world scientific discovery.

Looking to the future, we want to examine how our approach with \textit{HypoGen} generalizes to other scientific domains. Our evaluation focused on computer science, and it remains an open question how well the fine-tuning on one domain generalizes to another. We also plan to expand our dataset to fields such as astrophysics, biology, and materials science, where hypothesis generation could accelerate scientific discoveries in fundamentally different fields. This work aims to enable interdisciplinary AI teammates that collaborate with human experts on challenging scientific tasks \citep{Swanson2024}, with the overarching goal of democratising science.

\section*{Acknowledgments}

The authors are deeply grateful to Dr. Charles F. McMillan, whose encouragement to pursue bold ideas inspired this work, and we dedicate this study to him. We thank Microsoft Research and the Microsoft Accelerating Foundation Models Research program for their continuous support and for providing the OpenAI credits used to generate the \textit{HypoGen} outputs. We also thank the Oak Ridge Leadership Computing Facility for access to high-performance computing resources that supported this research.

\bibliography{colm2025_conference}
\bibliographystyle{colm2025_conference}

\appendix
\section{Appendix: Prompts used in the analysis}
\label{sec:appendix_extras}

\subsection{Abstract-Level Bit-Flip-Spark Prompt}
\begin{lstlisting}[breaklines=true, basicstyle=\small\ttfamily]
ABSTRACT_PROMPT = """
You are a highly advanced research assistant, specialized in reading scientific papers for hypothesis generation and identifying innovative ideas.

Before you begin, let's revisit the Bit-Flip concept with an example (BERT in NLP):

Bit: Traditional NLP models (RNNs, LSTMs) process text sequentially, limiting their ability to understand long-range dependencies and fully capture bidirectional context.
Flip: Instead, consider entire sentences at once, allowing context from both directions. This helps capture nuanced relationships among words.
Spark: Bidirectional context for NLP.

Bit-Flip Defined:
A Bit-Flip inverts a commonly held assumption, questioning existing constraints or reapplying techniques to new domains/scales. The "Bit" is the prevailing belief, and the "Flip" is the counterargument.

Guidance for Analysis (Abstract-Level):
1. **Bit (Technical Insight)**:
   - Provide at least **two** sentences clearly stating the status quo or conventional approach.
   - Highlight the **limitation** or **problem** it creates.
   - Include **enough detail** so it is self-contained and does not rely on additional context from elsewhere.

2. **Flip (Innovation)**:
   - Provide at least **two** sentences describing the **novel approach** or perspective.
   - Explain the **method** or **technique** that enables this change.
   - Include **enough detail** so the Flip is understandable on its own.

3. **Spark (Core Summary)**:
   - A concise **4-6 word** phrase capturing the core idea.

Now, consider this research abstract:
{abstract}

Your task: Identify the Bit, Flip, and Spark from the abstract in a **detailed** manner:

- **Bit**: at least two sentences, with sufficient detail about the conventional approach and its limitation.
- **Flip**: at least two sentences, describing the new approach or perspective with enough detail to understand the main technique.
- **Spark**: a concise 4-6 word summary of the core idea.

Follow these rules:
- Do not cite the paper itself or its authors.
- Instead of saying "We/I introduced an idea", just say "An idea was introduced...".

Return ONLY the JSON object in **this exact format** (no extra text):
\\{{
  "Bit": "Technical limitation or conventional approach, in at least two sentences",
  "Flip": "Innovative approach or solution, in at least two sentences",
  "Spark": "4-6 word summary"
\\}}
"""
\end{lstlisting}

\subsection{Chain-of-Reasoning Prompt}
\begin{lstlisting}[breaklines=true, basicstyle=\small\ttfamily]
NOTEBOOK_PROMPT = """
You are a highly advanced computer scientist with extraordinary ability in scientific hypothesis generation.

You are given:
- A pre-identified "Bit"(the conventional assumption or limitation)
- A pre-identified "Flip"(the innovative approach or solution)
- The full text of the paper.

Please provide the Scientist's Ideation Notebook to obtain the intellectual process that went from Bit to Flip. In other words, how did the Bit go to the Flip? The goal is to model the intellectual process of a scientist in a comprehensive cycle of analysis, summarizing, exploration, reassessment, reflection, backtracing, and iteration to develop a well-considered thinking process as they understand how to go from the Bit to the Flip.

This scientist_notebook should be detailed enough to write the paper and must include a mix of interrogative and reflective output. It needs to include questions that probe the process alongside reflective answers that elaborate on methodological details, as well as experimental observations and results that emerged during hypothesis testing. It should also include a few additional important questions regarding experimental design, data analysis, and validation, and contain a reflection that highlights a breakthrough insight akin to a Eureka moment, without stating that you experienced one.
It needs to be written in first person singular and follow these rules:

Rules:
- Use scientific language.
- Ensure that the scientist_notebook includes explicit questions that probe your reasoning process, clearly interwoven with your reflective responses.
- Use only evidence from the paper text, don't quote it but rephrase it in a more concise form.
- Be very specific and clear about methodological details. Integrate technical and methodological details in a reflective style that explains and justifies each inquiry.
- You can use parts of the provided Bit or Flip, but do not incorporate their text verbatim.
- Do not use generic phrases such as The Bit suggests...—always use the actual content of the Bit.
- Only citations referenced in the paper are allowed. Do not make up citations.
- Keep the notebook concise and with great logical flow, with a maximum of ten relatively short sentences, optimally. Do not use overlong sentences.
- When specific methods or models are mentioned, incorporate further context provided in the paper text to strengthen your analysis.
- Include discussion of experimental results and additional probing questions related to experimental design, data analysis, and validation.
- Do not mention you experienced an "Eureka!" moment, but provide a question or reflection that clearly highlights a breakthrough insight akin to a turning point.
- The output needs to be in continuous flow, for example, no bullet points or numbered lists.
- Have good grammar, syntax and punctuation.

Bit: {bit}
Flip: {flip}

Paper text:
{paper_text}

Return ONLY the JSON below (no other text):
{{
  "notebook": scientist_notebook
}}
"""

\appendix
\section{Prompts Used in Experiments}

\subsection{Abstract-Level Analysis Prompt}
\begin{lstlisting}[breaklines=true, basicstyle=\small\ttfamily]
ABSTRACT_PROMPT = """
...
"""
\end{lstlisting}

\subsection{Evaluation Prompt}
\begin{lstlisting}[breaklines=true, basicstyle=\small\ttfamily]
prompt = f"""
I need you to evaluate two different proposed solutions to a problem. I'll provide the problem statement and two options (A and B), each with a "Spark" (the core idea) and a "Chain of Reasoning" (detailed explanation).

PROBLEM:
{problem}

OPTION A:
Spark: {option_a['spark']}
Chain of Reasoning: {option_a['chain']}

OPTION B:
Spark: {option_b['spark']}
Chain of Reasoning: {option_b['chain']}

Please evaluate both options on two dimensions:
1. NOVELTY: Which option presents a more novel/creative approach?
2. FEASIBILITY: Which option is more practical and likely to succeed?

Then determine an OVERALL WINNER based on your holistic assessment.

Format your response exactly like this:
MORE NOVEL: [A, B, or NONE]
MORE FEASIBLE: [A, B, or NONE]
OVERALL WINNER: [A, B, or TIE]

Then provide a brief explanation of your reasoning (2-3 sentences).
"""
\end{lstlisting}

\clearpage

\subsection{Model Generations}

\subsection{Original Bit}
\begin{tcolorbox}[
  colback=gray!5,    % Very light gray background
  colframe=gray!50,   % Medium gray border
  boxrule=1pt,
  arc=2mm,
  auto outer arc,
  after=\vspace{-3pt}  % Reduce space after
]
Conventional information retrieval methods rely on large labeled datasets for fine-tuning, which can be expensive or impractical to obtain. Moreover, even when such datasets are available, their effectiveness rapidly decreases in new domains due to distributional shifts.
\end{tcolorbox}

\subsubsection{\texttt{LLaMA-8B-1shot}}
\begin{tcolorbox}[
  colback=blue!5,    % Very light blue background
  colframe=blue!40,   % Medium blue border
  boxrule=1pt,
  arc=2mm,
  auto outer arc,
  before=\vspace{-3pt},  % Reduce space before
  after=\vspace{-3pt}    % Reduce space after
]
I started by considering the limitations of conventional information retrieval methods. They heavily rely on large labeled datasets, which can be expensive or impractical to obtain in some cases. Moreover, even when such datasets are available, their effectiveness rapidly decreases in new domains due to distributional shifts. This led me to wonder: how might I address these challenges without breaking the bank or requiring extensive data annotation? I explored whether utilizing self-supervised learning techniques could help bridge this gap. By generating pseudo-labels from unlabeled data and using them for fine-tuning, I asked: could we leverage weakly supervised or even unsupervised methods to adapt conventional retrieval models more efficiently? Another question that arose was: might we tap into the strengths of pre-trained language models by leveraging their rich contextual representations? This led me to ponder whether augmenting these representations with domain-specific knowledge graphs could enhance the model's robustness and generalizability across diverse domains. As I delved deeper, I began to consider the potential benefits of meta-learning. By training a model on multiple tasks or datasets simultaneously, it might learn more transferable and flexible representations that can adapt quickly to novel environments. This sparked another question: how would we evaluate the performance of such models in real-world settings? Upon further reflection, I realized that addressing these challenges may involve a multidisciplinary approach. By integrating insights from natural language processing (NLP), graph neural networks (GNNs), and meta-learning, we could potentially develop more adaptive and domain-agnostic retrieval methods.
\end{tcolorbox}

\subsubsection{\texttt{LLaMA-8B-FT}}
\begin{tcolorbox}[
  colback=green!5,    % Very light green background
  colframe=green!40,   % Medium green border
  boxrule=1pt,
  arc=2mm,
  auto outer arc,
  before=\vspace{-3pt},  % Reduce space before
  after=\vspace{-3pt}    % Reduce space after
]
I began by asking whether large labeled datasets were necessary to fine-tune a retrieval model, and I found that in many cases, they are not. I tested a method that learns to rank from unlabeled data and confirmed that it outperformed standard fine-tuning approaches on several benchmarks. I then questioned how to ensure that the model generalizes well in new domains, and I introduced a domain-specific pre-training stage to adapt the model to different distributions. I asked how to measure the impact of this pre-training, so I performed extensive experiments that showed consistent improvements across diverse domains. I also examined how to validate that the model's rankings were robust, and I measured the correlation between the predicted rankings and true relevance scores. I found that the pre-training stage effectively bridged the gap between different distributions, and I wondered if further hyperparameter tuning would yield even better performance. I tested different data augmentation strategies and found that they further enhanced the model's adaptability. I reflected on whether this approach could scale to larger datasets, and I confirmed that it maintained its effectiveness even with millions of unlabeled examples. Finally, I concluded that domain-specific pre-training on unlabeled data can greatly reduce reliance on labeled data and enable more efficient retrieval in new domains.  
\end{tcolorbox}

\subsubsection{\texttt{R1-distilled-LLaMA-1shot}}
\begin{tcolorbox}[
  colback=orange!5,    % Very light orange background
  colframe=orange!40,   % Medium orange border
  boxrule=1pt,
  arc=2mm,
  auto outer arc,
  before=\vspace{-3pt},  % Reduce space before
  after=\vspace{-3pt}    % Reduce space after
]
I recognized the challenge of costly labeled datasets in information retrieval. I hypothesized that pre-trained models could offer a foundation for new domains without extensive fine-tuning. I explored transfer learning strategies to adapt these models to specific tasks, testing domain adaptation techniques for better generalization. I also considered self-supervised learning to leverage unlabeled data and enhance robustness across distributions. Through empirical validation, I demonstrated that these approaches significantly reduced dependency on large datasets while maintaining performance.
\end{tcolorbox}

\subsubsection{\texttt{R1-distilled-LLaMA-FT}}
\begin{tcolorbox}[
  colback=purple!5,    % Very light purple background
  colframe=purple!40,   % Medium purple border
  boxrule=1pt,
  arc=2mm,
  auto outer arc,
  before=\vspace{-3pt},  % Reduce space before
  after=\vspace{-3pt}    % Reduce space after
]
I first asked how to train retrieval models without relying on large labeled datasets, and I tested a few approaches that only use unlabeled text. I hypothesized that self-supervised methods could offer a more efficient training path, so I designed a pipeline that first pre-trained a retrieval model on unlabeled text and then used a small labeled subset to refine it. I observed that this approach improved retrieval performance over methods that solely used labeled data, especially in domains with limited labeled samples. To validate this conclusion, I conducted multiple experiments on diverse domains, measuring retrieval accuracy and efficiency. I noticed that the pre-training phase was crucial for capturing general patterns, while the refinement step fine-tuned the model to better match specific labels. I then asked if this approach would generalize across different datasets and found consistent improvements across multiple benchmark sets. I tested the model on a domain with a large labeled set and observed that it performed as well as the fine-tuned model, indicating that the self-supervised pre-training was sufficient. I also questioned whether the refinement step was truly necessary and found that omitting it degraded retrieval performance. In analyzing the data, I realized that the small labeled subset was vital for aligning the model with the target tasks, confirming the importance of both pre-training and refinement. I then considered whether the approach would remain robust if the labeled subset was too small, and the results indicated that it still performed adequately. Ultimately, I concluded that combining self-supervised pre-training with a small labeled refinement step is an effective and efficient method for training retrieval models in diverse domains. and efficient method for training retrieval models in diverse domains.
\end{tcolorbox}

\clearpage
\section{o3-mini Evaluation}
\label{sec:o3-mini}
\begin{figure}[h!]  % h! gives highest priority to placing it "here"
    \centering
    \includegraphics[width=\linewidth]{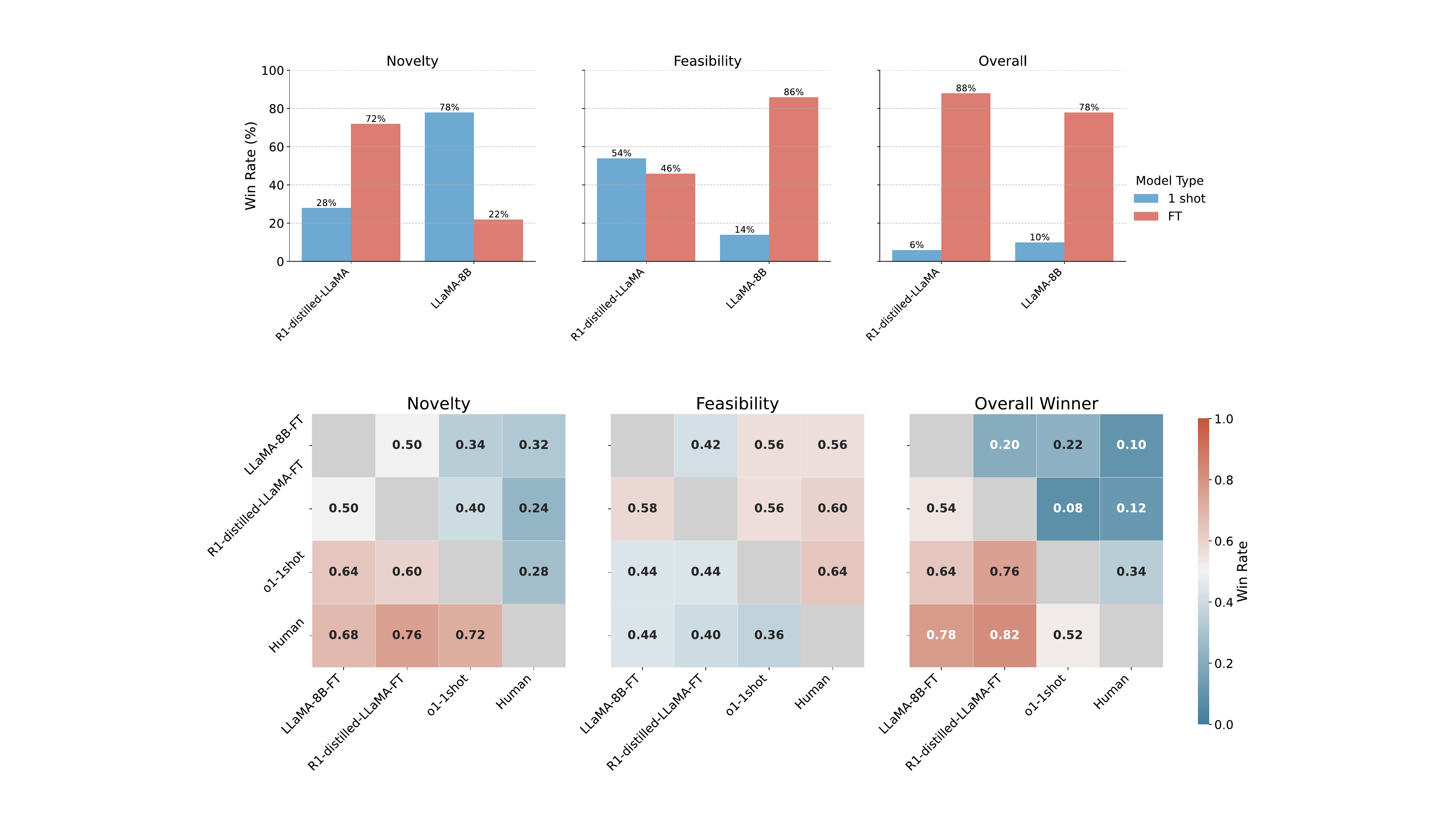}
    \caption{Comparative analysis of the quality of generated hypotheses across nine experiments as evaluated by an LLM Judge \texttt{o3-mini}.}
    \label{fig:results-o3}
\end{figure}

\end{document}